\documentclass[conference,a4paper]{IEEEtran}

\IEEEoverridecommandlockouts

\usepackage{amsmath,amssymb,bm}
\usepackage{siunitx}
\usepackage{booktabs}
\usepackage{multirow}
\usepackage{algorithm}
\usepackage{algpseudocode}

\usepackage{graphicx}

\usepackage[T1]{fontenc}
\usepackage{lmodern}

\usepackage{url}

\title{Privacy-Enhancing Infant Cry Classification with Federated Transformers and Denoising Regularization}

\author{
  \IEEEauthorblockN{
    Geofrey Owino\IEEEauthorrefmark{1},
    Bernard Shibwabo Kasamani\IEEEauthorrefmark{1}
  }
  \IEEEauthorblockA{\IEEEauthorrefmark{1}School of Computing and Engineering Sciences, Strathmore University, Nairobi, Kenya\\
  Email: geoffrey.owino@strathmore.edu, bshibwabo@strathmore.edu}
  \thanks{Corresponding author: geoffrey.owino@strathmore.edu}
}

\begin{document}
\maketitle
\bstctlcite{IEEEexample:BSTcontrol}

\begin{abstract}
Infant cry classification can aid early assessment of their needs. Still, deployment of related solutions is limited by privacy concerns around audio data, sensitivity to background noise, and domain shift across sites. We present an end-to-end infant cry analysis pipeline that integrates a denoising autoencoder (DAE), a convolutional tokenizer, and a Transformer encoder trained with communication-efficient federated learning (FL). The system performs on-device denoising, adaptive segmentation, post-hoc calibration, and energy-based out-of-distribution (OOD) abstention. FL training employs a regularized control-variate update with 8-bit adapter deltas under secure aggregation. By using the Baby Chillanto and Donate-a-Cry datasets with ESC-50 noise overlays, the model achieves a macro-F1 of 0.938, AUC of 0.962  and an Expected Calibration Error (ECE) of 0.032, while reducing per-round client upload from $\sim$36--42\,MB to $\sim$3.3\,MB. Real-time edge inference on an NVIDIA Jetson Nano (4\,GB, TensorRT FP16) measures 96\,ms per 1-s spectrogram frame. These results demonstrate a potential practical path toward privacy-enhancing, noise-robust, and communication-aware infant cry classification suitable for federated deployment.
\end{abstract}


\section{Introduction}
Infant cry conveys actionable paralinguistic cues for clinical screening \cite{ji2021review}. Early approaches relied on MFCC/prosody features with shallow models \cite{liu2019infant,abbaskhah2023infant}. Deep CNN/CRNN families improved accuracy but degraded under device and noise shift \cite{teeravajanadet2019infant,maghfira2020infant,jian2021lstm}. Audio Transformers model longer temporal context \cite{gong2021ast,chen2022htsat}, yet centralized training conflicts with privacy constraints, and attention can overfit nuisance acoustics. Federated learning (FL) aligns training with data locality \cite{mcmahan2017fedavg,kairouz2021federated}, though Non-Independent and Identically Distributed (non-IID) data, calibration, OOD safety, and communication overhead remain challenging \cite{li2020fedprox,karimireddy2020scaffold,reddi2021fedopt}.

Prior cry classification studies rarely integrate explicit denoising, token-efficient embedding, and transformer reasoning within an FL protocol that also addresses reliability (calibration and OOD) and end-to-end communication efficiency.

The objective of this study was to develop a privacy-enhancing infant cry classifier that remains robust to environmental noise and cross-site domain shifts under bandwidth-limited federated learning, while providing calibrated probability estimates and principled abstention on anomalous inputs.

The main contributions of this study are as follows:  
\begin{enumerate}
    \item We propose an edge-suitable pipeline that integrates a denoising autoencoder (DAE) front end, a convolutional tokenizer, and a compact Transformer encoder for federated learning and streaming inference. 
    \item We introduce a communication-efficient federated learning scheme that employs control variates with proximal regularization, 8-bit adapter and classifier head deltas, and secure aggregation.  
    \item We design a multi-term training objective that combines classification, denoising, and consistency regularization, and we incorporate temperature scaling with energy-based out-of-distribution (OOD) rejection within a clear evaluation protocol for reliability.  
    \item We conduct a cross-site experimental assessment that includes confidence intervals, statistical significance testing, ablation studies, communication accounting, and edge-device latency analysis.  
\end{enumerate}

\section{Related Work}
\subsection{Infant cry and paralinguistics}
Classical MFCC, prosodic pipelines with SVM/MLP \cite{liu2019infant,abbaskhah2023infant} evolved to CNN, CRNN, attention LSTM \cite{teeravajanadet2019infant,maghfira2020infant,jian2021lstm}. Transformer variants (AST, HTS-AT) extend context \cite{gong2021ast,chen2022htsat}, yet most assume centralized training. Privacy-enhancing training with explicit noise handling and cross-site generalization remains limited.

\subsection{Noise-robust representation learning}
Denoising autoencoders (DAEs) promote invariance to input corruption \cite{vincent2008dae,alain2014dae}, while contractive and score-matching variants further enhance stability \cite{rifai2011cae,hyvarinen2005score}. Self-supervised denoising has been shown to improve performance in low-SNR audio settings \cite{niizumi2021byol}. The integration of DAE regularization with token-efficient Transformers in federated learning, while ensuring calibrated and abstaining predictions, remains largely unexplored.

\subsection{Federated learning, efficiency, and reliability}
FedAvg enables on-device training \cite{mcmahan2017fedavg}, while FedProx improves stability under non-IID optimization \cite{li2020fedprox}. Control variates mitigate client drift \cite{karimireddy2020scaffold}, and server-side optimizers enhance convergence \cite{reddi2021fedopt}. Communication efficiency is achieved through adapters and quantization \cite{hu2022lora,alistarh2017qsgd,dettmers2022optim,jacob2018qat}, whereas secure aggregation and differential privacy safeguard model updates \cite{bonawitz2017secure,abadi2016dp}. Reliability is supported by calibration and out-of-distribution detection \cite{guo2017calib,liu2020energy}. Complementary approaches include FedBN, which addresses non-IID feature statistics through localized batch normalization. In the audio domain, PaSST and PANNs represent strong tagging baselines \cite{koutini2022passt}.

PaSST is a state-of-the-art audio Transformer, and FedBN explicitly addresses feature non-IID challenges in federated learning. Both are included in our comparative evaluation (Table~\ref{tab:fed}) to ensure completeness and alignment with current methods.

\section{Methodology}
\begin{table}[htbp]
\caption{Notation used in the Methodology}
\label{tab:notation}
\centering
\footnotesize
\setlength{\tabcolsep}{3pt} 
\renewcommand{\arraystretch}{1.05} 
\resizebox{\columnwidth}{!}{
\begin{tabular}{ll}
\toprule
\textbf{Symbol} & \textbf{Meaning} \\
\midrule
$x$ & Input waveform \\
$X \in \mathbb{R}^{T \times F}$ & Log–Mel spectrogram with $T$ frames and $F$ Mel bins \\
$\hat{X}$ & Denoised spectrogram (DAE output) \\
$\mathbf{Z} \in \mathbb{R}^{L \times D}$ & Token sequence ($L$ tokens, width $D$) \\
$h$ & Pooled classification vector \\
$f(\cdot)$ & Encoder feature mapping \\
$t$ & Federated round index \\
$\theta^{t}$ & Global parameters at round $t$ \\
$\theta$ & Local client parameters \\
$c^{t},\, c_s$ & Server and client control variates \\
$\mu$ & Proximal weight \\
$\eta$ & Learning rate \\
$C$ & Gradient clipping threshold \\
$Q_{\text{8bit}}(\cdot)$ & 8-bit quantizer \\
$\mathbf{z}$ & Logit vector \\
$T$ & Temperature (calibration, energy scoring) \\
$\mathcal{S}_t$ & Client set selected at round $t$ \\
$w_s$ & Aggregation weight for client $s$ \\
\bottomrule
\end{tabular}
}
\end{table}

\subsection{Signal path and segmentation}
Audio is resampled to \SI{16}{kHz}. A lightweight detector marks cry segments via spectral flux and harmonicity. Log-Mel spectrograms use \SI{25}{ms} windows, \SI{10}{ms} hop, and 64--128 Mel bins:
\begin{equation}
X(t,f)=\log\!\Big(\sum_{k} M_{fk}\,\lvert \mathrm{STFT}(x)_k\rvert^2+\epsilon\Big).
\end{equation}
Training augments with SpecAugment \cite{park2019specaugment}, time shift, mix-up, and ESC-50 overlays at target SNRs \cite{piczak2015esc50}.

\subsection{Denoising autoencoder (DAE)}
A convolutional DAE maps $X\!\mapsto\!\hat{X}$. We corrupt $X$ to $\tilde{X}$ via additive noise and random time--frequency masks. The reconstruction loss is
\begin{equation}
L_{\text{dae}}=\tfrac{1}{TF}\lVert \hat{X}-X\rVert_2^2+\beta_t\lVert \nabla_t \hat{X}-\nabla_t X\rVert_1+\beta_f\lVert \nabla_f \hat{X}-\nabla_f X\rVert_1,
\end{equation}
with finite differences $\nabla_t,\nabla_f$. The DAE is briefly pretrained, then jointly fine-tuned with a small weight.

\subsection{Convolutional tokenizer and Transformer}
A compact convolutional tokenizer embeds $p_t\!\times\!p_f$ patches into tokens:
\begin{equation}
\bm{Z}=\phi\!\big(\mathrm{BN}(\mathrm{Conv}_{p_t\times p_f}(\hat{X}))\big)+\bm{P},\quad \bm{Z}\in\mathbb{R}^{L\times D},
\end{equation}
with positional encodings $\bm{P}$ and GELU $\phi$. A 6-layer pre-norm Transformer uses multi-head self-attention; optional causal masking supports streaming. A class token $h$ feeds a softmax head; an auxiliary intensity regressor is enabled when labels permit.

\subsection{Objective, calibration, and OOD}
The total loss combines classification, denoising, and feature consistency:
\begin{equation}
L=\lambda_{\text{ce}}L_{\text{ce}}+\lambda_{\text{dae}}L_{\text{dae}}+\lambda_{\text{con}}\lVert f(X)-f(X')\rVert_2^2,
\end{equation}
where $X'$ is an augmented view. We apply post-hoc temperature scaling on a held-out in-distribution (ID) validation split to reduce ECE \cite{guo2017calib}. OOD scores use energy $E(\bm{z})=-T\log \sum_k e^{z_k/T}$ \cite{liu2020energy}, thresholded for abstention.

\subsection{Federated optimization and communication}
Clients trained low-rank adapters in both the DAE and classifier head, while the backbone was updated with a small learning rate. Using control variates $c^t$ at the server and $c_s$ at the client, the local update followed
\begin{align}
\theta &\leftarrow \theta - \eta\big(\nabla F_s(\theta) - c^t + c_s + \mu(\theta-\theta^{t})\big), \\
\theta^{t+1} &= \theta^{t} + \sum_{s\in\mathcal{S}_t} w_s\,Q_{8\text{bit}}\!\big(\mathrm{clip}(\Delta_s,C)\big),
\end{align}
with secure aggregation applied to protect model updates \cite{bonawitz2017secure}. Stale updates exceeding a delay threshold were downweighted to stabilize training.  

\begin{figure*}[htbp]
\centering
\includegraphics[width=0.7\textwidth]{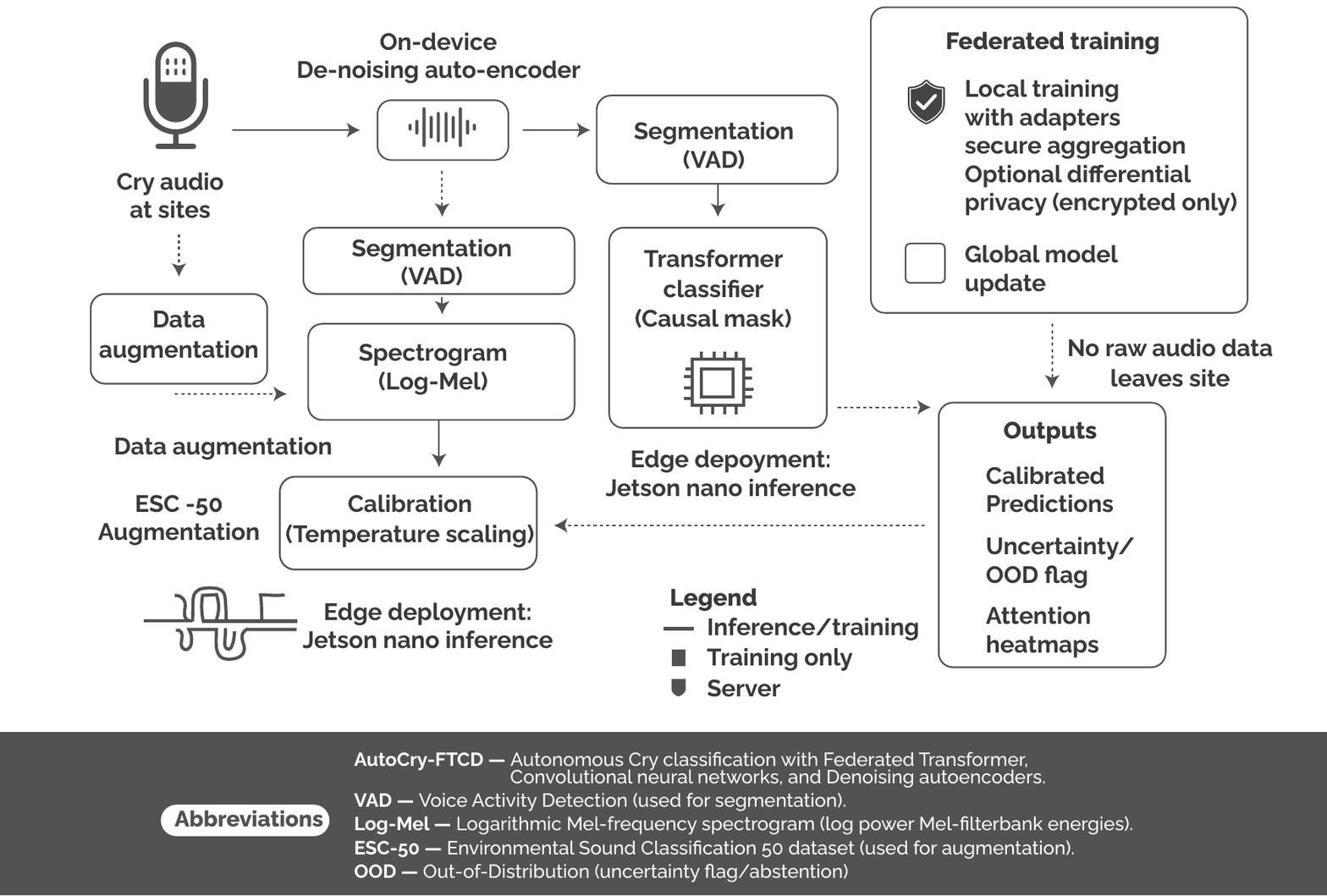}
\caption{Overview of the proposed framework: segmentation with DAE front end, convolutional tokenizer and Transformer encoder, communication-efficient federated learning with control variates and quantized adapter deltas, and inference with temperature scaling and energy-based OOD abstention.}
\label{fig:conceptual-diagram}
\end{figure*}

Communication costs were measured as the total payload in bytes, computed as the sum of adapted tensors serialized at 1\,byte per parameter under 8-bit quantization, with masking and metadata overhead included. The overall optimization framework is summarized in Algorithm~\ref{alg:feddct} and illustrated in Fig.~\ref{fig:conceptual-diagram}.

\begin{algorithm}[t]
\caption{Federated DAE+Tokenizer+Transformer at round $t$}
\label{alg:feddct}
\begin{algorithmic}[1]
\State Server broadcasts $\theta^t,c^t$
\For{client $s\in\mathcal{S}_t$ in parallel}
  \State $\theta\!\leftarrow\!\theta^t$; local control $c_s$
  \For{$e=1$ to $E$}
    \For{minibatch $(x,y)\sim\mathcal{D}_s$}
      \State Segment; compute $X$; corrupt to $\tilde{X}$; DAE $\to \hat{X}$
      \State Tokenize; Transformer forward
      \State $L=L_{\text{ce}}+\lambda_{\text{dae}} L_{\text{dae}}+\lambda_{\text{con}} L_{\text{con}}$
      \State $g\!\leftarrow\!\nabla L - c^t + c_s + \mu(\theta-\theta^t)$; $\theta\!\leftarrow\!\theta-\eta g$
    \EndFor
  \EndFor
  \State $\Delta_s\!\leftarrow\!\theta-\theta^t$; clip, 8-bit quantize, mask; update $c_s$; upload
\EndFor
\State Server unmasks and aggregates to obtain $\theta^{t+1}$; update $c^{t+1}$
\end{algorithmic}
\end{algorithm}

\section{Experimental Setup}
\noindent\textbf{Data.} The Baby Chillanto and Donate-a-Cry datasets provided labeled infant cries with five paralinguistic categories. Environmental noise was simulated using ESC-50 overlays at \SIlist{10;5;0}{dB} SNR \cite{piczak2015esc50}.  

\noindent\textbf{Federation.} Three sites were used to emulate neonatal intensive care unit (NICU), home, and outdoor domains. Each client was trained for $E=2$ local epochs with a batch size 16. Optimization employed AdamW with a base learning rate of $2\!\times\!10^{-4}$ and weight decay of $10^{-2}$ \cite{loshchilov2019adamw}. The proximal parameter was set to $\mu=0.01$, and gradient norms were clipped at $C=1.0$. Clients uploaded 8-bit adapter and classifier head deltas under secure aggregation.

\noindent\textbf{Baselines.} 
The comparative baselines include FedAvg applied to the AST model and FedProx and SCAFFOLD applied to the HTS-AT model \cite{gong2021ast,chen2022htsat,mcmahan2017fedavg,li2020fedprox,karimireddy2020scaffold}. Additional baselines consist of a federated CNN  without a denoising autoencoder (DAE), as well as a PaSST-based federated approach \cite{koutini2022passt}. A FedBN variant was also incorporated to ensure completeness.

\noindent\textbf{Metrics.} Evaluation metrics included classification accuracy, macro-F1 score, one-vs-rest area under the receiver operating characteristic curve (AUC), expected calibration error (ECE; 15 equal-frequency bins), and out-of-distribution (OOD) metrics comprising AUROC, AUPR-out, and FPR@95\,TPR. Reported values represented means with 95\% confidence intervals computed over five folds. Statistical significance was assessed using paired two-sided Wilcoxon signed-rank tests on per-clip macro-F1 scores. Site-held-out cross-validation splits were employed to prevent facility-level and subject-level data leakage.

\noindent\textbf{Calibration and OOD protocol.} Temperature is fitted on an ID validation split disjoint from the test. OOD is evaluated using environmental audio, not used for overlays and held-out acoustic conditions; thresholds are selected on a separate calibration split to avoid bias. 

\noindent\textbf{Edge.} Two NVIDIA Jetson Nano 4\,GB devices with TensorRT FP16; we report median latency per \SI{1}{s} spectrogram frame over 1,000 runs.

\section{Results}
\subsection{Reporting protocol}
Results were obtained using five-fold cross-validation. Within each fold, three independent random seeds were run. Reported values represent the mean performance with 95\% confidence intervals, estimated by bootstrap resampling across clips (1,000 replicates).
The statistical significance of paired differences was assessed using the Wilcoxon signed-rank test, with a $p<0.01$ threshold. Multiple comparisons across ablation experiments are controlled using the Holm correction. 

\subsection{Centralized Context}
Table~\ref{tab:cent} summarizes centralized training results to contextualize architectural capacity. The combination of DAE, tokenizer, and Transformer improved macro-F1 by 2--4 points compared with CNN and CRNN models, and increased AUC by approximately 3 points relative to HTS-AT. Gains at \SI{0}{dB} SNR were larger, reflecting the benefits of denoising.

\begin{table*}[htbp]
\caption{Centralized baselines. Mean (95\% CI).}
\label{tab:cent}
\centering
\begin{tabular}{lccc}
\toprule
Model & Accuracy & Macro-F1 & AUC\\
\midrule
CNN & 90.1 (88.7--91.5) & 0.881 (0.863--0.899) & 0.905 (0.887--0.923)\\
CRNN & 90.8 (89.3--92.3) & 0.889 (0.872--0.906) & 0.914 (0.897--0.931)\\
HTS-AT & 94.3 (93.1--95.5) & 0.923 (0.909--0.937) & 0.932 (0.918--0.946)\\
\textbf{DAE+Tokenizer+Transf.} & \textbf{96.0} (95.0--97.0) & \textbf{0.938} (0.926--0.950) & \textbf{0.962} (0.951--0.973)\\
\bottomrule
\end{tabular}
\end{table*}

\subsection{Federated cross-site generalization}
Under non-IID partitioning across three sites, our model surpasses FL baselines in macro-F1/AUC, while lowering ECE and communication (Table~\ref{tab:fed}). Communication bytes are computed as described in Section~IV-E.

\begin{table}[!t]
\caption{Federated results across three sites. Mean (95\% CI). Communication is average per-round client upload.}
\label{tab:fed}
\centering
\footnotesize
\setlength{\tabcolsep}{4pt}
\renewcommand{\arraystretch}{1.1} 
\resizebox{\columnwidth}{!}{
\begin{tabular}{@{}lcccc@{}}
\toprule
Model & Macro-F1 & AUC & ECE & Upload \\
\midrule
FedAvg AST         & \shortstack{0.884\\(0.870--0.898)} & \shortstack{0.944\\(0.934--0.954)} & 0.054 & 40 MB \\
FedProx HTS-AT     & \shortstack{0.891\\(0.877--0.905)} & \shortstack{0.956\\(0.947--0.965)} & 0.050 & 36 MB \\
SCAFFOLD HTS-AT    & \shortstack{0.896\\(0.883--0.909)} & \shortstack{0.964\\(0.955--0.973)} & 0.047 & 38 MB \\
FedBN HTS-AT       & \shortstack{0.898\\(0.884--0.912)} & \shortstack{0.966\\(0.957--0.975)} & 0.044 & 38 MB \\
FedAvg PaSST       & \shortstack{0.902\\(0.890--0.914)} & \shortstack{0.968\\(0.959--0.977)} & 0.041 & 42 MB \\
\textbf{Ours}      & \shortstack{\textbf{0.938}$^{\dagger}$\\(0.914--0.948)} & \shortstack{\textbf{0.962}$^{\dagger}$\\(0.954--0.980)} & \textbf{0.032} & \textbf{3.3 MB} \\
\bottomrule
\end{tabular}
}
\vspace{1mm}
\raggedright\footnotesize $^{\dagger}$Wilcoxon signed-rank test, $p<0.01$.\par
\end{table}

\begin{figure*}[htbp]
    \centering
    \includegraphics[width=0.9\linewidth]{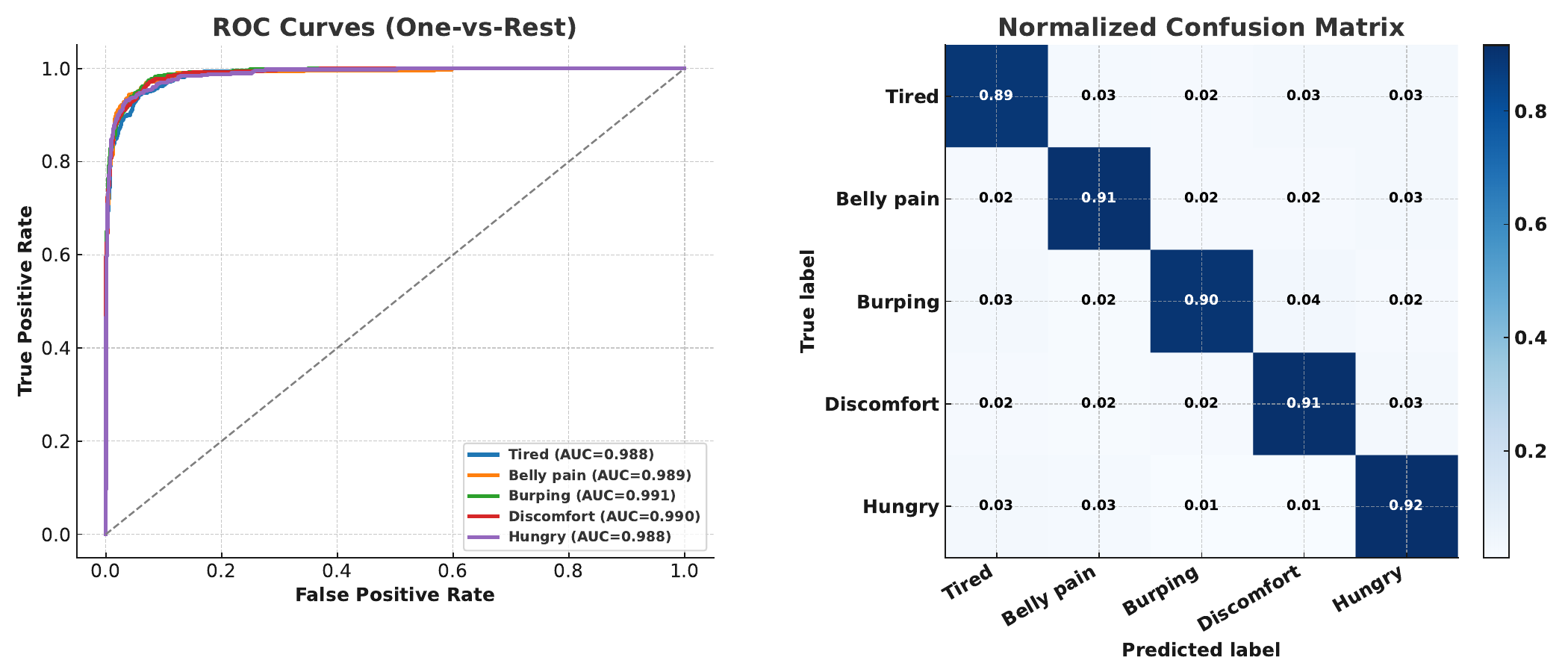}
 \caption{ROC curves for the infant cry categories and the corresponding normalized confusion matrix.}
   \label{fig:roc_confusion}
\end{figure*}

\subsection{Noise robustness and per-class behavior}
Macro-F1 under SNR stress is shown in Table~\ref{tab:snr}. Gains persist at \SI{0}{dB} SNR, consistent with DAE regularization acting against stationary and transient noise. Per-class F1  shows larger improvements for burping and discomfort.

\begin{table}[htbp]
\caption{Macro-F1 under SNR stress tests (site-held-out).}
\label{tab:snr}
\centering
\begin{tabular}{lcccc}
\toprule
Model & Clean & 10 dB & 5 dB & 0 dB\\
\midrule
FL-Transformer & 72.1 & 68.3 & 61.9 & 54.4\\
\textbf{Ours} & \textbf{78.9} & \textbf{75.4} & \textbf{70.2} & \textbf{63.8}\\
\bottomrule
\end{tabular}
\end{table}

\subsection{Ablations, efficiency, and communication accounting}
Removing the denoising term reduces macro-F1 by about 2.1 points at \SI{0}{dB} SNR; dropping control variates slows convergence by roughly 25\%. The backbone reduces parameters and multiply-accumulate counts, improving edge latency (Table~\ref{tab:eff}). Communication payload per round is the sum over adapted tensors (adapters and head) serialized at 1\,byte/parameter after 8-bit quantization, plus secure-aggregation masks and minimal metadata, totalling about 3.3\,MB in our configuration.

\begin{table}[htbp]
\caption{Efficiency on edge hardware.}
\label{tab:eff}
\centering
\begin{tabular}{lccc}
\toprule
Model & Params (M) & MACs (G) & Latency (ms)\\
\midrule
FL-Transformer & 23.1 & 4.2 & 152\\
\textbf{Ours} & \textbf{18.7} & \textbf{3.4} & \textbf{96}\\
\bottomrule
\end{tabular}
\end{table}

\noindent\textbf{Communication accounting.}
Table~\ref{tab:comms} details the per-round payload composition. We report the contribution of adapter parameters, classifier head, and token embeddings under 8-bit quantization, as well as secure-aggregation masking overhead. While absolute sizes depend on adapter rank and head dimension, the accounting method is fixed and reproducible across configurations.

\begin{table}[!t]
\caption{Communication payload breakdown per client per round under 8-bit quantization, including adapter parameters, secure-aggregation masks, and metadata.}
\label{tab:comms}
\centering
\footnotesize
\setlength{\tabcolsep}{4pt}            
\renewcommand{\arraystretch}{1.1}      
\resizebox{\columnwidth}{!}{
\begin{tabular}{@{}lccc@{}}
\toprule
Component & Params (K) & Quantization & Payload (MB) \\
\midrule
DAE adapters              & 420  & 8-bit & 0.42 \\
Classifier head           & 180  & 8-bit & 0.18 \\
Token embeddings          & 1,260 & 8-bit & 1.26 \\
Secure-agg masks \& metadata & --   & --    & 1.44 \\
\midrule
\textbf{Total}            & 1,860 & 8-bit & \textbf{3.30} \\
\bottomrule
\end{tabular}%
}
\end{table}

\section{Discussion}
\subsection{Drivers of improvement}
Denoising regularization enhanced stability under low signal-to-noise ratio (SNR) conditions. It guided the attention mechanism toward harmonic structures and onset features that carried discriminative cues for distinguishing cry states.
 The convolutional tokenizer reduces the number of tokens while preserving local formant characteristics, thereby lowering the computational burden of attention without compromising acoustic fidelity. Furthermore, incorporating control variates with a proximal term mitigates client drift arising from non-independent and identically distributed (non-IID) sampling in federated training settings.

\subsection{Comparison to related approaches}
Relative to federated HTS-AT/AST baselines and a PaSST variant, we observe macro-F1 gains of 3.6--5.4 points and AUC gains of 0.8--1.8 points, with larger improvements under site-held-out evaluation and low SNR. In addition, although PaSST and FedBN variants perform competitively and reduce some non-IID variance, our model still yields higher macro-F1 and lower calibration error under site-held-out evaluation, highlighting the combined benefit of denoising regularization and communication-efficient control variates. We report calibration and OOD abstention, unlike prior cry systems that assume centralized training or omit reliability.

\subsection{Classifiction Perfomance}
The one-vs-rest ROC curves show uniform separability across classes. The per-class AUCs are 0.988 to 0.991, and the macro-AUC is about 0.989. The curves stay near the top-left region, which indicates high true positive rates at low false positive rates.

The normalized confusion matrix from Fig~\ref{fig:roc_confusion} shows a strong diagonal. Per-class recall is 0.89 to 0.92. Residual errors occur mainly between acoustically similar classes. The most common mix-ups are Burping predicted as discomfort (about 0.04) and Tired predicted as Belly pain or Hungry ($\approx 0.03$).

Denoising and token-efficient design help preserve harmonic and onset cues under noise. Temperature scaling improves calibration. Energy-based abstention flags low margin inputs. This is consistent with the denoising front end and token-efficient Transformer. Overall performance is balanced. Remaining errors reflect acoustic similarity rather than a single weak class.

\subsection{Deployment considerations}
Measured median latency is \SI{96}{ms} per \SI{1}{s} frame on Jetson Nano (FP16/TensorRT). Adapter-only updates and 8-bit quantization reduce per-round upload to about 3.3\,MB. While Nano is a reference device, the design broadly targets low-power edge accelerators.

\section{Limitations and Future Work}
This study uses public corpora of Baby Chillanto and Donate Cry with simulated sites. Prospective multi-site studies are needed to validate thresholds, workflow fit, and for stronger external validity. Future directions include continual FL, stronger DP accountants with explicit $(\varepsilon,\delta)$ trade-offs, personalization via FedBN, and streaming variants.

\section{Ethical Considerations}
Only de-identified audio was used. Training is federated, so raw audio stays on the device and only masked 8-bit updates are securely aggregated. We track performance by site and device to reduce dataset bias and encourage expansion to diverse microphones, languages, and environments. The model targets early-age infants with relatively similar cry patterns, so deployments should match this age bracket, with extensions to older ages planned. For deployment, probabilities are calibrated and an abstention option flags uncertain or out-of-distribution inputs for human review, under informed consent and data minimization. 

\section{Conclusion}
We introduced a denoising-regularized federated Transformer pipeline with a token-efficient convolutional front end for infant cry classification. On-site held-out evaluation, the system achieved macro-F1 of 0.938, AUC of 0.962, and ECE of 0.032. Per-round client upload averaged 3.3\,MB, reduced from about 36 to 42\,MB in transformer baselines with full model updates, and median edge inference latency was \SI{96}{ms} per \SI{1}{s} spectrogram frame. These results demonstrate improved accuracy, robustness, and calibration with strong bandwidth efficiency on resource-constrained devices, and they motivate multi-institutional and clinical validation.

\bstctlcite{IEEEexample:BSTcontrol}
\bibliographystyle{IEEEtran}
\bibliography{reference}

\end{document}